\pdfoutput=1

\documentclass[11pt]{article}

\usepackage[]{acl}

\usepackage{times}
\usepackage{latexsym}
\usepackage{booktabs}
\usepackage[T1]{fontenc}
\usepackage{multirow}
\usepackage[utf8]{inputenc}

\usepackage{microtype}
\usepackage{subfig}
\usepackage{xspace}

\usepackage{inconsolata}

\usepackage{graphicx}
\usepackage{amsmath, amssymb}

\newcommand{\evaluator}{\textsc{ProbEval}\xspace}
\newcommand{\gpt}{GPT2}

\title{Every Answer Matters: Evaluating Commonsense \\ with Probabilistic Measures}

\author{Qi Cheng$^{1}$
\quad Michael Boratko$^{2}$\Thanks{Now at Google DeepMind, Eightfold.ai, Thorn, and Meta respectively.} 
\quad Pranay Kumar Yelugam$^{2 *}$
\quad Tim O'Gorman$^{2 *}$  \\
\bf \quad Nalini Singh$^{2 *}$
\quad Andrew McCallum$^{2}$ 
\quad Xiang Lorraine Li$^{1}$\\
$^1$University of Pittsburgh,
$^2$University of Massachusetts Amherst \\ 
\texttt{\{qic69, xianglli\}@pitt.edu}}

\begin{document}
\maketitle

\begin{abstract}
Large language models have demonstrated impressive performance on commonsense tasks; however, these tasks are often posed as multiple-choice questions, allowing models to exploit systematic biases~\cite{li2021systematic}. Commonsense is also inherently probabilistic with multiple correct answers. The purpose of “boiling water” could be making tea, cooking but also could be killing germs. Existing tasks do not capture the probabilistic nature of common sense. To this end, we present commonsense frame completion (CFC), a new generative task that evaluates common sense via multiple open-ended generations \footnote{Dataset and PROBEVAL available at: \url{https://github.com/qxc101/PROBEVAL_CFC/}}. We also propose a method of probabilistic evaluation that strongly correlates with human judgments. Humans drastically outperform strong language model baselines on our dataset, indicating this approach is both a challenging and useful evaluation of machine common sense.
\end{abstract}

\section{Introduction}
Most existing commonsense evaluations utilize multiple-choice question-answering (MCQA) tasks~\cite{talmor2018commonsenseqa, sap2019socialiqa, huang2019cosmos, bhagavatula2020abductive, Feng2022GenericTR}. This format offers a limited view of common sense. MCQA tasks simplify the problem with unrealistically small answer sets, and making the options challenging is difficult~\cite{zellers2018swagaf, hellaswag}. More crucially, common sense is \emph{implicit} – understanding assumptions that are unspoken precisely because they are common knowledge.  MCQA, by evaluating common sense explicitly, fails to capture a model's ability to utilize this knowledge in unprompted, generative contexts. MCQA is also fundamentally at-odds with the fact that common sense is inherently probabilistic, and should be evaluated as such.

\begin{figure}[t]
    \centering
    \includegraphics[width=\columnwidth]{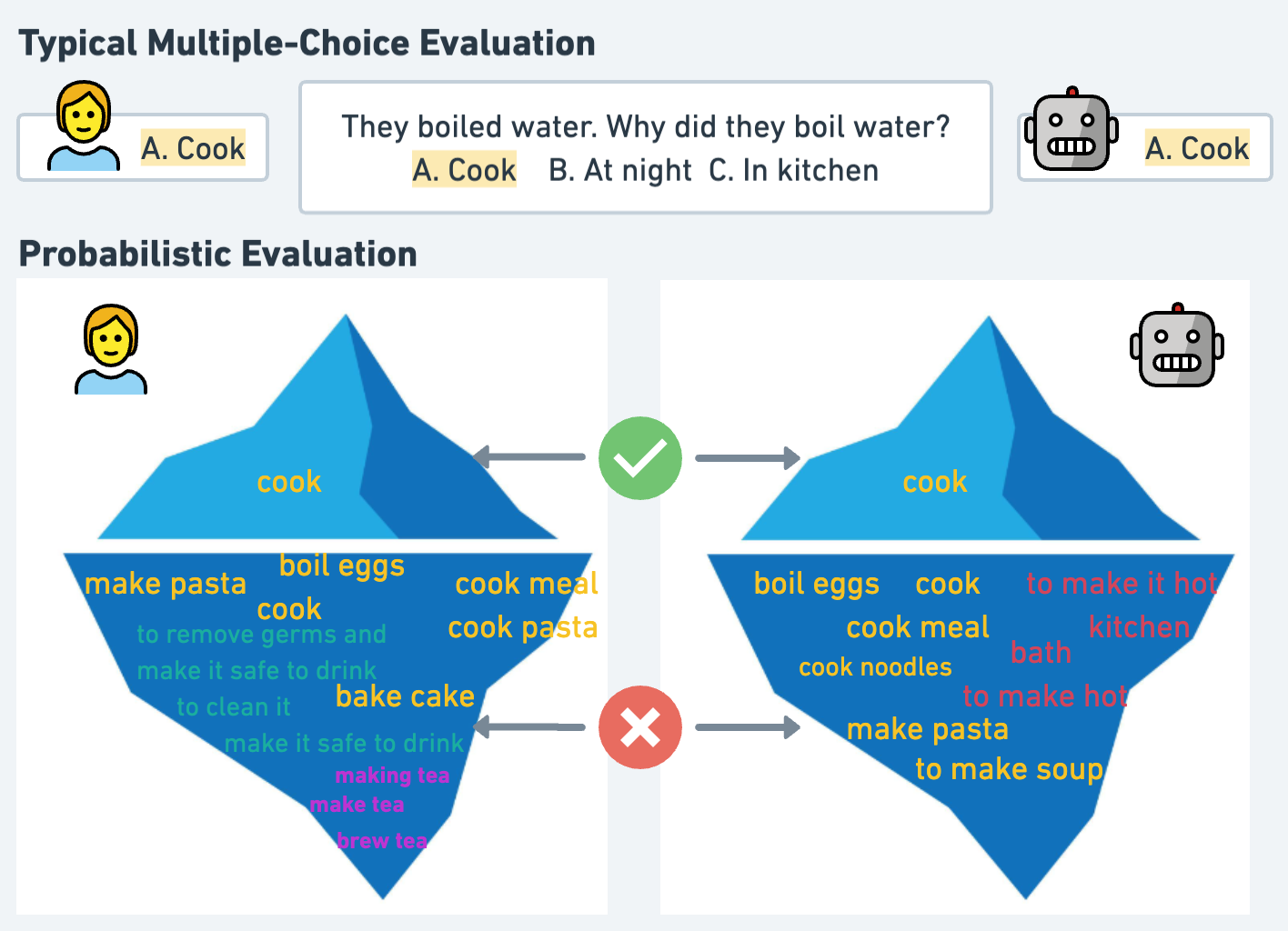}
    \caption{Typical evaluations only compare human and model performance for their top choices (top). We propose to evaluate multiple plausible answer choices by clustering similar answers, denoted by color, to form categorical distributions,
    and evaluating probabilistically (bottom). This more accurately captures the probabilistic nature of common sense, and allows us to provide a more nuanced analysis of model capabilities.
    }
    \label{fig:intro_fig}
\end{figure}

In order to avoid the issues in MCQA, many recent benchmarks have proposed generative commonsense evaluations.~\citep{Lin2020CommonGenAC, chen2023say}. While generative evaluation avoids the difficulty of generating hard negatives, it does not reflect the fact that there are often multiple correct answers, especially for commonsense questions. For example, in the phrase "they boiled the water", we can infer using our common sense that the most likely reason for this action is cooking or making tea. However, people in areas with limited clean water access may view this as a way to remove germs and ensure it’s safe to drink. This aspect is, unfortunately, frequently overlooked during the benchmark creation process. To ensure that the model can serve diverse populations, it is important to gather multiple responses. By focusing on collecting implicit information from larger population, this provides a more accurate evaluation of common sense required in real-world settings.

In this setting, it is crucial to address the commonsense questions with multiple correct answers.
While some previous works have proposed a clustering-and-ranking evaluation~\citep{boratko2020protoqa}, we demonstrate that such an approach can lack nuance.
Instead, we embrace the probabilistic nature of common sense and evaluate the model's ability to capture a probability distribution over answer clusters sampled from the population.
This ensures that the model not only captures common sense, but does so in a calibrated way.

To this end, we propose commonsense frame completion (CFC), a task focused on inferring missing information in a context sentence. This emphasizes the \textit{implicit} nature of common sense in a given context and is tightly connected to downstream applications, such as home assistants, where inferring such information about a user's query is key. In CFC, the questions are generated by identifying missing information from a given context sentence about daily scenarios. For each context-question pair, we collected a large number of diverse answers from human annotators. 

In order to evaluate multiple responses effectively, we additionally propose a new evaluation method. Responses from different individuals often vary, and common sense is typically defined as the knowledge shared by nearly all people. To make every answer count, we consider the annotators' answers from a \textit{probabilistic} viewpoint and evaluate models using probabilistic measures~\citep{Moss2018-MOSPK-2, Pavese2020-PAVPKI, CHATER2006287}. Specifically, given raw string answers, we cluster them and consider the categorical distribution over these clusters based on the number of answers contained in each. We propose automated clustering and alignment mechanisms which allow models to be evaluated directly by comparing the KL divergence between distributions. The task format and evaluation metric are shown in Figure~\ref{fig:intro_fig}. We evaluate the proposed metric with two datasets, CFC and ProtoQA~\citep{boratko2020protoqa}, and show high correlations with human judgments for both datasets. 

Finally, we report multiple LLMs' performance on CFC measured by the proposed probabilistic metrics. We identify a large performance gap between existing large models and humans, indicating the limitations of current LLMs.
\section{Related Work}
\paragraph{Commonsense Evaluation}
Creating commonsense benchmarks to evaluate model performance is a long-standing research topic~\cite{winogrande, Lin2020CommonGenAC, sap-etal-2019-social, zhou2020evaluating}. However, most benchmarks are created using a multiple-choice selection paradigm, which is simpler to evaluate but misaligned with the real-world use-case of commonsense knowledge, and most egregiously ignores the existence of multiple correct answers. We are not the first ones to gather multiple human answers to facilitate robust evaluations, however. \citet{aydin2014crowdsourcing} and \citet{boratko2020protoqa} also collected multiple human responses for each question to get aggregated human ground-truth answer sets.

Our work differs from these due to our emphasis on commonsense as \emph{implicit} and \emph{probabilistic}. We don't treat each answer equally; rather, we aim to match the answer distribution given by human responses. For this purpose, we propose a novel probabilistic evaluation for open-ended generation tasks with multiple correct answers. A similar probabilistic evaluation was studied from a language model generation point of view~\cite{pillutla2021mauve}. They proposed a KL-based evaluation to measure language model generations, while our focus is on the implicit answer distribution.

Among the previous work, our proposed dataset CFC is most similar to ProtoQA~\cite{boratko2020protoqa} as they share a similar task format (one question, multiple answers). However, there are several key differences. CFC is designed to uncover implicit commonsense information in various contexts, making it a more generalized variant of ProtoQA. Consequently, the question format in CFC is significantly more diverse than that of ProtoQA. In addition, the number of answers collected for CFC is rigorously justified through the Neyman-Pearson lemma, guaranteeing a sufficient and representative sample.

\begin{table*}[]
\centering
\small
\begin{tabular}{@{}l|l|l@{}}
\textbf{Missing Slot} & \textbf{Definition}                                                                                                           & \textbf{Examples}                                                                                                                            \\ \midrule
Arg0         & Who/what does the event?                                                                                             & \begin{tabular}[c]{@{}l@{}}Sentence: putting cheese on the pizza. Arg0?\\ Answers: person, cook\end{tabular}                        \\ \midrule
Purpose      & What is the goal for doing the event?                                                                                & \begin{tabular}[c]{@{}l@{}}Sentence: putting cheese on the pizza. Purpose?\\ Answers: get nutrition, stop being hungry\end{tabular} \\ \midrule
Instrument   & What kind of tools are used to accomplish the event?                                                                 & \begin{tabular}[c]{@{}l@{}}Sentence: putting cheese on the pizza. Instrument?\\ Answers: hands, spoon\end{tabular}                  \\ \midrule
Time         & \begin{tabular}[c]{@{}l@{}}What is a particular time (time of day, season, etc.)\\ for doing the event?\end{tabular} & \begin{tabular}[c]{@{}l@{}}Sentence: putting cheese on the pizza. Time?\\ Answers: lunch time, dinner time\end{tabular}             \\ \midrule
Location     & Where would the event usually happen?                                                                                & \begin{tabular}[c]{@{}l@{}}Sentence: putting cheese on the pizza. Location?\\ Answers: kitchen, restaurant\end{tabular}            
\end{tabular}

\caption{Examples for different missing slot types in CFC}
\label{table:dataset}
\end{table*}
\paragraph{Commonsense as Probabilistic Knowledge}
In most knowledge evaluation benchmarks, commonsense knowledge is defined as absolute facts~\citep{bian2023chatgpt, chen2023say}. We relax this absolute intersection between human knowledge using a probabilistic approach. The probabilistic notion of knowledge is well supported. \citet{Moss2018-MOSPK-2} stated human beliefs or credences, inherently probabilistic, should be regarded as legitimate forms of knowledge. The growing trend towards building probabilistic models in cognitive studies reinforces the idea of human cognition and memory functioning as probabilistic processors \citep{CHATER2006287}. This collective body of work supports the view that common sense serves as probabilistic assumption instead of definite judgements in human mind.

\section{CFC Task Description}

In this section we motivate and describe the task of "commonsense frame completion" (CFC). We aim to create a task which evaluates implicit common sense with multiple correct answers. Given a direction such as ``put the water on the burner to boil”, it is common sense which allows us to understand that the water is likely in a kettle and not simply dumped on the burner.  Unlike factual question answering tasks, there is no single correct answer. In this example, the water could be placed in a ``kettle”, ``pot”, ``cup'', or ``glass”, although the former answers are more probable. 

It is necessary for any machine learning model which claims to capture common sense to correctly predict all the possible answers and have some sense of the distribution over the implicit information. To assess a model's ability in this regard, we view the context sentence as a structured semantic frame, identify a missing slot, and ask the model to provide a distribution of potential slot fillers.

\section{Dataset Creation and Analysis}
We now describe the dataset creation process of CFC. We first need to collect reasonable context sentences which contain natural element of common sense. CommonGen~\citep{Lin2020CommonGenAC} is a commonsense dataset which contains many short sentences describing basic information about daily life, and so we choose this dataset as the source for potential context sentences.

Given a short sentence, we then identify implicit information. To this end, we perform semantic parsing on the sentence, and identify missing slots. We use AMR~\citep{Banarescu2013AbstractMR} for semantic parsing based on its ability to provide a rich representation of the sentence with a pre-defined fixed schema for the predicate roles. If a predicate is found, AMR parsing will match it to a schema and fill in the values for any identified slots. Any slots marked with \texttt{amr-unkown} indicate potential items of missing information, enabling us to obtain human annotations for the missing slot values.

We uniformly sampled 63,788 sentences from the CommonGen dev dataset, and parsed them using the AMR parser from \citet{cai-lam-2020-amr}, generating 228,170 pairs of context questions with missing slots. From this, we randomly sampled 101 (sentence, missing slot) pairs for crowd workers to annotate, such that we had a balanced distribution of missing slot types, as detailed in Section \ref{analysis:answer_numbers}. We present the context sentence and missing slot to crowdworkers, who were also provided with training examples and descriptions of the meaning of each slot type (see Table~\ref{table:dataset}). The number of answers is chosen such that the resulting answer distribution is stable (see Section \ref{analysis:answer_numbers}). Each element of the raw dataset therefore includes a context sentence, missing slot value, and a collection of slot fillers.

\subsection{Probability Distribution}
\label{dataset:probability}
In an open-ended task where multiple humans are asked to provide answers as raw strings of text there are a multitude of answers which may essentially capture the same underlying idea. Ultimately we are not interested in the different variations of the surface form, but rather in capturing the essence of the underlying concept. In the boiling water example, we may want to treat "kettle" and "teapot" as though they were representative of the same general concept. As originally proposed in \citet{boratko2020protoqa}, we consider \emph{clustering} the responses, converting a set of answer strings into a categorical distribution over answer clusters, where the probability of obtaining an answer from a given cluster is proportional to the number of answer strings contained within it. We explore both manual clustering and automated clustering methods (see Section~\ref{sec:Probabilistic Evaluation}).

\subsection{Analysis}
\paragraph{Number of Answers}
\label{analysis:answer_numbers}
The number of potential slot fillers might be very large, and we want to ensure we sample enough to approximate the true distribution over answer concepts. An essential question is how many samples are enough to approximate the true distribution with reasonable error rate? This is a classic problem in statistics, for which the Neyman-Pearson lemma proves that the uniformly most powerful test is to consider the KL divergence $D_\text{KL}(g \Vert f) = \sum_x g(x) \log \frac{g(x)}{f(x)}$ where $g$ is the empirical distribution and $f$ is the true distribution~\cite{harremoes2012information}. The recent work from \citet{mardia2020concentration} showed that this can be bounded by the following equation

\small
\begin{equation*}
    \mathbb{P}(D_{KL}(g_{n,k} \Vert f)\geq \epsilon) \leq e^{-n \epsilon}\left [ \frac{3c_1}{c_2} \sum _{i=0}^{k-2} K_{i-1} (\frac{e\sqrt{n}}{2\pi})^i\right ]
\end{equation*}

\normalsize

where $c_1$ and $c_2$ are constant values, $n$ is the number of samples, and $k$ is the number of categories in the categorical distribution.

\begin{figure}[t]
\centering
\small
\subfloat[]
{\includegraphics[width=0.238\textwidth]{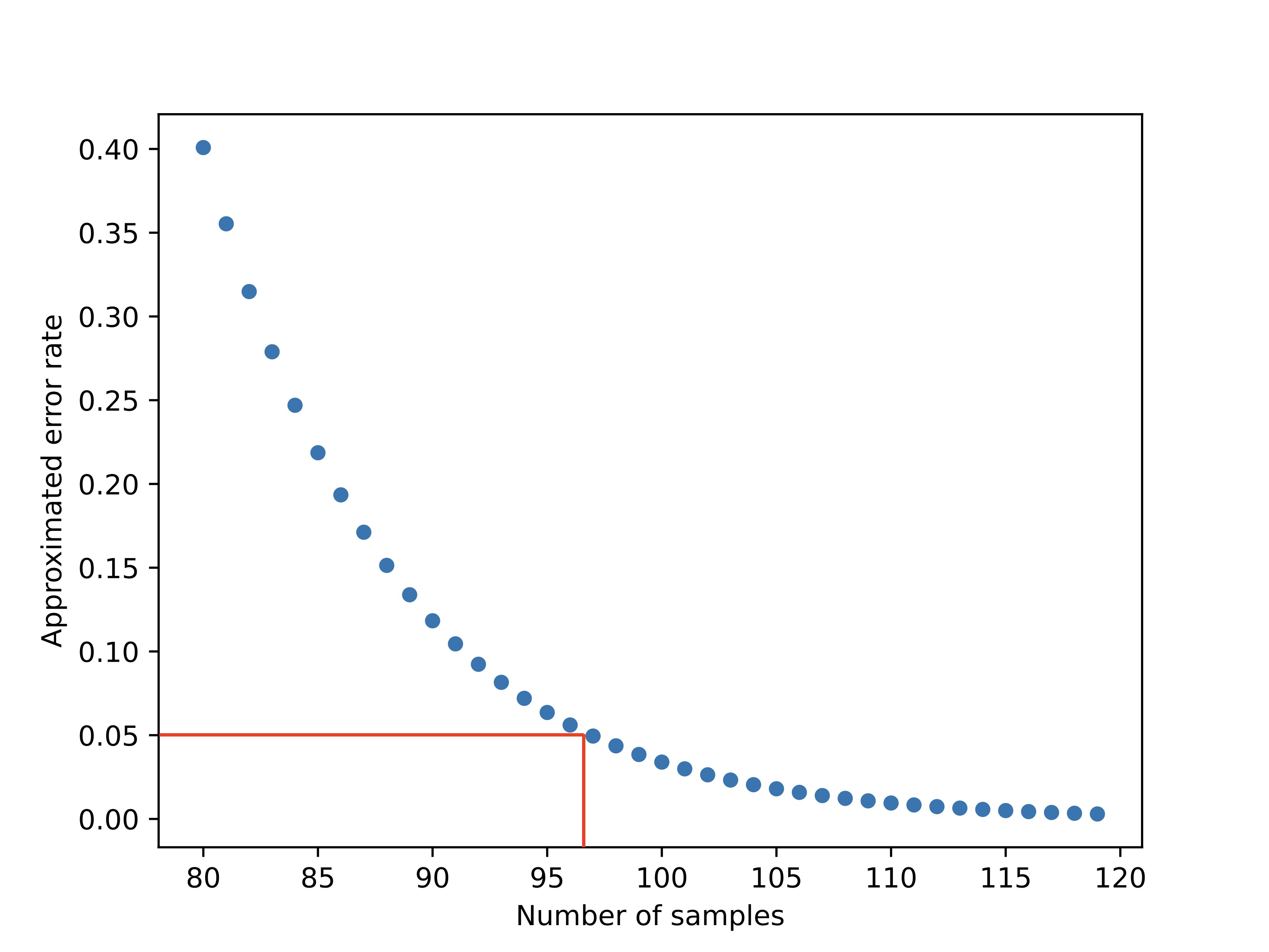}}
\subfloat[]
{\includegraphics[width=0.238\textwidth]{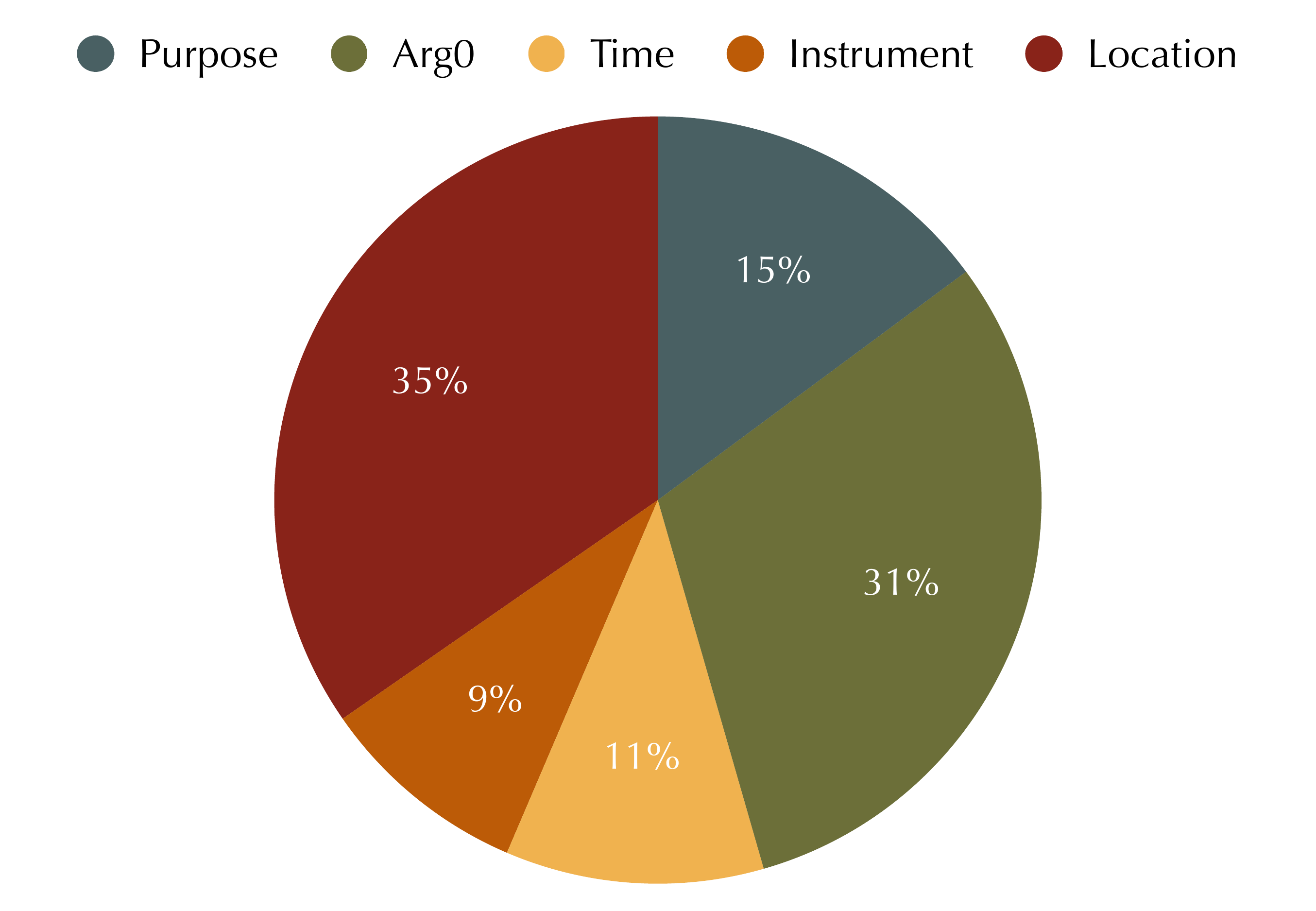}}
\caption{(a) The relationship between the number of examples (x-axis), and the approximation error rate (y-axis) calculated using the bound from~\citet{mardia2020concentration}. (b) Question type distribution in CFC.}

\label{analysis:types}
\end{figure}

In our setting, we manually clustered 50 questions, and found that the number of categories is not more than 8. To get a bound on the number of answers we should collect, we set $\epsilon=0.2$, $k=8$, and solve $e^{-n \epsilon}\left [ \frac{3c_1}{c_2} \sum _{i=0}^{k-2} K_{i-1} (\frac{e\sqrt{n}}{2\pi})^i\right ]$ for $n$.
Figure~\ref{analysis:types} (a) shows the value of this bound on the $y$-axis for increasing numbers of samples $n$ on the $x$-axis. As we can see from the graph, for $100$ samples,  the error rate is less than $0.05$, allowing us to approximate the true answer distribution with $95\%$ confidence if there are fewer than 8 categories in the categorical distribution.

\paragraph{Question Types}

We collected $101$ (context, missing slot) pairs, and obtained 100 slot fillers for each from crowdworkers, resulting in 10,100 annotations overall. The data collection page we used on Amazon MTurk is shown in Fig \ref{fig:AMT}\footnote{The annotators are paid 0.15 per answer, and they are all anonymous English speakers who are based in the US.}. We create a dev set with 55 examples and a test set with 46 examples. The distribution of missing slot types are shown in Figure \ref{analysis:types} (b). Each question type is associated with a different type of commonsense reasoning, e.g time represents temporal commonsense reasoning. 
\section{Probabilistic Evaluation}
\label{sec:Probabilistic Evaluation}
In this section, we detail the method of evaluating multiple correct answers. As we relaxed commonsense to be probabilistic knowledge, a rigorous probabilistic evaluation is required; however the task is presented (both to humans and models) as a generative question answering task. Therefore, we need a way to compare two large sets of answer strings. We will start by how human evaluators may compare the similarity of there sets of answers and then describe the various ways by which this process can be automated. 

\subsection{General Framework}
\label{sec:General Framework}
\begin{figure}[ht]
\centering
\includegraphics[width=0.45\textwidth]{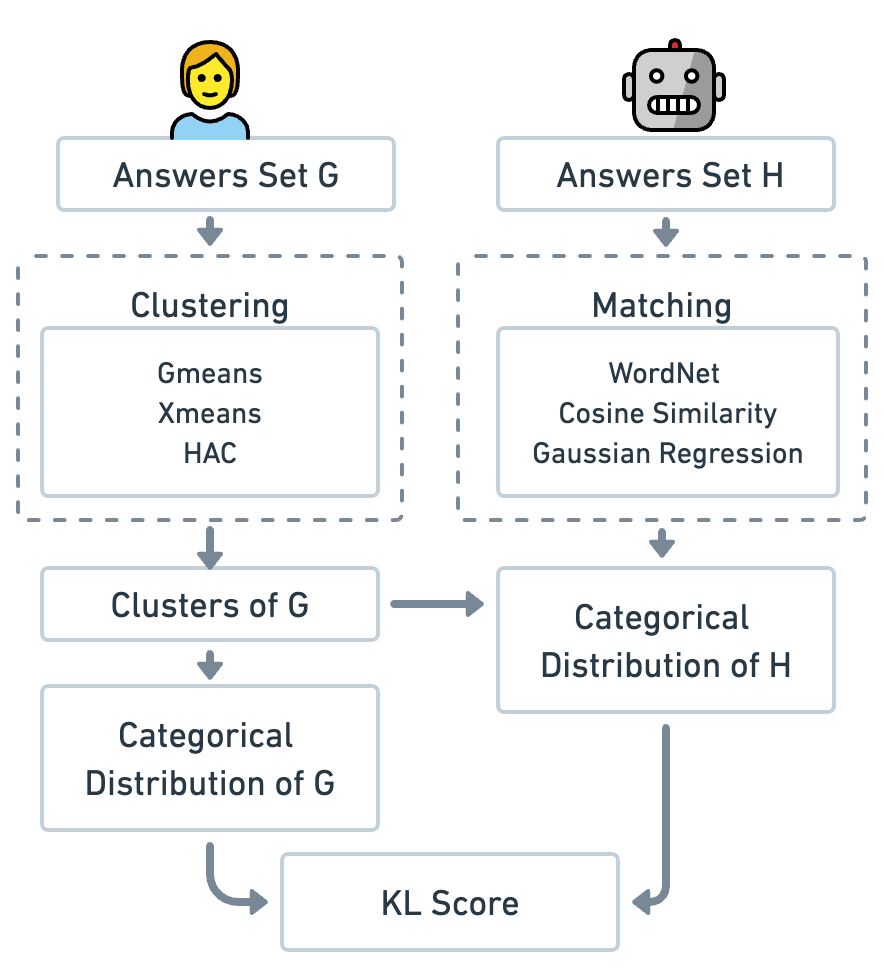}
\caption{Given a question, the ground truth answer set $G$ from human, and the model predicted answers set $H$. The first step is to form clusters of $G$ with concept-level meaningful clusters. The clusters could form categorical distribution of $G$. We then would match each answer in $H$ into created clusters of $G$. After matching, we would form a categorical distribution of $H$. Finally, we calculate the KL score between these two distributions. }
\label{fig:evaluator_step}
\end{figure}
Given a question, the ground truth answer set $\mathbf{G}$ and the model generated answers $\mathbf{H}$, the goal is to evaluate the similarity between these two answer sets. This is a difficult task even for a human, especially if the answer sets are large and diverse, however, we cares more about \emph{concepts} being captured rather than unique surface forms. So we start with clustering the answer strings in $\mathbf{G}$ to form meaningful concept level clusters.\footnote{When clustering, a new category "wrong" is added to the answer set to account for the wrong answers for a question. These will then be discarded prior to model evaluation.} We could then match the answers in $\mathbf{H}$ to the proposed ground-truth clusters in $\mathbf G$. Upon having the clusters, we could define categorical probability distributions over the clusters, $P_g$ and $P_h$, where the probability assigned to a given cluster is proportional to the number of answer strings assigned to it.%
\footnote{To eliminate zero probabilities, we use Laplace smoothing on all categories before calculating the probabilities, --- adding one dummy answer to all categories.}
Finally, the similarity between $\mathbf{G}$ and $\mathbf{H}$ can be inferred by calculating the KL divergence of the two distributions, $D_\text{KL}(\hat{P}_g||\hat{P}_h)$. The overall evaluation framework is depicted in Figure \ref{fig:evaluator_step}.

\subsection{Automatic Evaluation}
\label{sec:automatic_eval}

Based on the general framework, we propose an automatic metric, \evaluator. The key steps are: 1. Embed ground-truth answers from $\mathbf G$ into vector space. 2. Automatically cluster the embeddings to obtain ground-truth clusters of $\mathbf G$. 3. Match elements of $\mathbf H$ to clusters of $\mathbf G$ by assignment function score. Each step presents a number of options, which we detail in the following sections.

\paragraph{Embedding}
\label{eval:embedding}
We first embed the discrete word tokens in $\mathbf G$ and $\mathbf H$ as word vectors. We experimented with various word embedding models, both without context(Word2Vec~\cite{word2vec}, GloVe~\cite{glove} and FastText~\cite{fasttex}) and with context (BERT~\cite{bert}, and RoBERTa~\cite{roberta}) We found FastText to perform best, and use it for all future embedding components. For answers with multiple tokens, we use the average of the FastText embeddings to represent those answer.

\paragraph{Clustering}
\label{eval:cluster}
Given the vector representation of the word answers, we experimented with various clustering algorithms including X-means~\cite{pelleg2000x}, G-means~\cite{zhao2008g} and hierarchical agglomerative clustering (HAC)~\cite{murtagh2014ward}. We used the implementation from pyclustering~\cite{Novikov2019}. The parameters used by these clustering algorithms are treated as hyper-parameters and are tuned based on the correlation score as we discuss in section \ref{sec:val_eval}.

\paragraph{Matching}
\label{eval:matching}
Given the predicted answers, we aim to match the answers to one or multiple ground truth answer clusters. This was also a requirement for ProtoQA~\citep{boratko2020protoqa}, so we leverage the best-performing WordNet matching function from there. We also have embeddings for our answers, we consider embedding-based similarity matching functions. We train a Gaussian regression model for each cluster in the ground-truth answers. The regression takes one answer representation as input, and output is the label of whether the answer belongs to one particular cluster. We also experimented with cosine similarity as a alternative for Gaussin regression model. If an answer matches with multiple clusters we divide the weight evenly among all matching clusters.

\subsection{Validation of the CFC Evaluator}
\label{sec:val_eval}

In order to validate \evaluator's performance, we compared it with the human evaluation results on two generative datasets, ProtoQA~\citep{boratko2020protoqa} and CFC. A robust automatic evaluation method should align well with human judgment when given different model predicted answers. 

We started by taking a linear combination of the ground-truth distribution $\hat{P}_g$ and a uniform distribution $\hat{P}_u$ to create diverse distributions that interpolates between the ideal ground truth answers to random noise, details in Appendix \ref{sec:Diverse Sampling}. However, arguably, the most important area to assess the quality of the evaluator is around answers which are likely to be returned from a model. We thus extend the above distribution by taking a linear combination of the answer distributions of a given baseline model $\hat{P}_h$, the ground-truth distribution $\hat{P}_g$, and a uniform distribution $\hat{P}_u$, with most of the weight assigned to the answers from a baseline model. This method predominantly features model-generated answers and is defined by the equation: $p = z\hat{P}_h + w'_1\hat{P}_g + w'_2\hat{P}_u$, where $w'_1$ and $w'_2$ are coefficients obtained from uniform distributions, emphasizing the blend of model insights with controlled randomness. The intuition behind this equation is that we want to define a new distribution $p$ , which is a weighted combination of distributions $\hat{P}_h$, $\hat{P}_g$ and a uniform distribution $\hat{P}_u$. We want these weights to be independent but still place most of the weight on the predicted distribution $hat{P}_h$. We achieve this by first sampling $z$ in $(0.5, 1)$, and then sampling two other values $w_1$ and $w_2$ in $(0, 1)$. We then computer $w'_1 = \frac{w’_1(1 - z)}{w_1 + w_2} $, and $w'_2 = \frac{w’_2(1 - z)}{w_1 + w_2} $. The coefficients are normalized in this process so that $z + w’_1 + w’_2 = 1$.

\subsubsection{Dataset}
We experimented with two datasets, the first one being ProtoQA \citep{boratko2020protoqa}. Both ProtoQA and CFC dev sets have 100 ground-truth answers and 30 additional human responses that were collected to measure human performance. For each question, in addition to the 130 human responses, we also use the 300 generated answers from the fine-tuned GPT2 model. All of these answers are annotated by expert annotators with cluster matching to the ground-truth clusters (details in Section \ref{subsec:gold_eval}). We use the GPT2 answers as the prediction set, $\mathbf{H}$. We sample 50 answer sets for each question from $\mathbf{H}$ and $\mathbf{G}$ according to the sampling procedure mentioned above. ProtoQA dev has 102 questions and CFC dev has 55 questions.

\subsubsection{Gold Evaluator Annotation}
\label{subsec:gold_eval}

An essential component in validation is to get human annotations for both clustering and matching. We requested two expert annotators to cluster 100 crowd-worker answers into 8 to 10 clusters independently as ground truth clusters. The inner-annotator agreement reached 0.76 when measured by BLANC~\citep{recasens2011blanc}. For model prediction matching in CFC, we take one of the baseline model predictions and employ GPT4 as silver annotation for matching. Two human annotators then verified the GPT4 matching, and reached 0.94 BLANC agreement. For ProtoQA, we used the human-annotated cluster and matching results in the paper~\citep{boratko2020protoqa}

We use automatic clustering and matching to get $D_\text{KL}(\hat{P}_g||\hat{P}_h)$. We can also evaluate the KL for manual clustering and matching, as all answers in ProtoQA have been annotated by human experts with clusters and assignments. After getting the human and automatic KL values for various sampled answer sets, we use the Spearman correlation coefficients across questions to measure the alignment between automatic and human evaluation.

\begin{table}[]
\centering

\begin{tabular}{@{}cc|cc@{}}
\toprule
                                                                              & Clustering & ProtoQA        & CFC            \\ \midrule
\multirow{4}{*}{\begin{tabular}[c]{@{}c@{}}ProtoQA \\ Evaluator\end{tabular}} & Human    & 0.193          & 0.257          \\ \cmidrule(l){2-4} 
                                                                              & Gmeans   & 0.167          & 0.239          \\
                                                                              & Xmeans   & 0.190           & 0.252          \\
                                                                              & HAC      & 0.193          & 0.252          \\ \midrule
\multirow{4}{*}{\begin{tabular}[c]{@{}c@{}} \evaluator \end{tabular}}     & Human    & 0.752 & 0.788 \\ \cmidrule(l){2-4} 
                                                                              & Gmeans   & 0.681          & 0.721          \\
                                                                              & Xmeans   & 0.669          & \textbf{0.728}          \\
                                                                              & HAC      & \textbf{0.698}          & \textbf{0.728}          \\ \bottomrule
\end{tabular}
\caption{Average Spearman correlation of ProtoQA evaluator and \evaluator compared with gold scores for ProtoQA and CFC dev questions. 
All entries use WordNet as matching function. \evaluator achieved much higher correlation compared to baseline evaluators. } 
\label{tab:correlation-dev}
\end{table}

\subsubsection{Experiment Setup}

\paragraph{Baseline} To compare \evaluator with other metrics, we assess the correlation between ProtoQA evaluator and human annotation. ProtoQA Evaluator evaluates multiple answer output as well, and it evaluates model predictions via ranked list of answers. The higher the score it, the better model prediction is. To accommodate ProtoQA score, The correlation is measured between 1 - ProtoQA scores (MaxAnswer@10) and KL divergence with gold clustering/matching annotations.

\subsubsection{Results}

\begin{table}[]
\centering
\begin{tabular}{@{}ll|ll@{}}
\toprule
Clustering & Matching & \multicolumn{1}{c}{ProtoQA} & \multicolumn{1}{c}{CFC} \\ \midrule
Human      & Cosine   & 0.715                       & 0.552                   \\
Human      & GR       & 0.708                       & 0.565                   \\ \midrule
Gmeans     & Cosine   & 0.528                       & 0.561                   \\
Gmeans     & GR       & 0.525                       & 0.541                   \\
Xmeans     & Cosine   & 0.525                       & 0.503                   \\
Xmeans     & GR       & 0.512                       & 0.505                   \\
HAC        & Cosine   & 0.585                       & 0.558                   \\
HAC        & GR       & \textbf{0.593}             & \textbf{0.564}       \\ \bottomrule
\end{tabular}
\caption{Average Spearman correlation for \evaluator when compared with gold scores. GR stands for Gaussian Regression. Correlation is generally higher when human is involved in the evaluation process (top two rows), but fully automatic evaluator \evaluator still achieves decent correlation.
}
\label{tab:correlation-dev-auto}
\end{table}
Our findings, detailed in Table \ref{tab:correlation-dev} and \ref{tab:correlation-dev-auto} \footnote{Scores are averaged cross 10 runs for setting with human clustering. }, indicate strong correlations between human evaluation and \evaluator. Table \ref{tab:correlation-dev} illustrates the correlation when using WordNet as the matching function with various automatic clustering algorithms. \evaluator demonstrates a stronger correlation with human scores when compared to the baseline, ProtoQA evaluator. Table \ref{tab:correlation-dev-auto} presents the results by using other matching functions that are not implemented in the ProtoQA evaluator in combination with different clustering algorithms. The result showed that whenever humans are involved in the evaluation process, whether in clustering or matching, the correlation is consistently higher. However, it is worth noting that a fully automatic evaluator with HAC as the clustering algorithm and WordNet as the matching function achieves around 0.7 correlation score that is nearly as good as humans.

\paragraph{Efficiency Analysis} We also report the running time of \evaluator. When using pre-trained word embeddings to calculate the similarity score, \evaluator requires approximately 0.319 seconds per question. When using WordNet as the similarity score, \evaluator takes about 49.47 seconds per question. The longer evaluation time associated with WordNet is a direct result of its multi-step process. This process becomes particularly time-consuming when the model predictions are long, as it involves the tokenization of the sentence followed by matching each token to ground truth tokens. Our main goal, however, was to come up with an automated evaluation to allow for efficient exploration, and our proposed approach is still much more efficient than human evaluation, which could take up to 10 minutes to perform clustering alone for a single question. Furthermore, human evaluation is expensive and not always feasible. Therefore, PROBEVAL offers an efficient and automated metric for the probabilistic evaluation of diverse answers.

\subsection{Qualitative Analysis}
\subsubsection{Visualization}

\begin{figure}[h]
\centering
\small
\subfloat[ProtoQA Score on CFC]
{\includegraphics[width=0.238\textwidth]{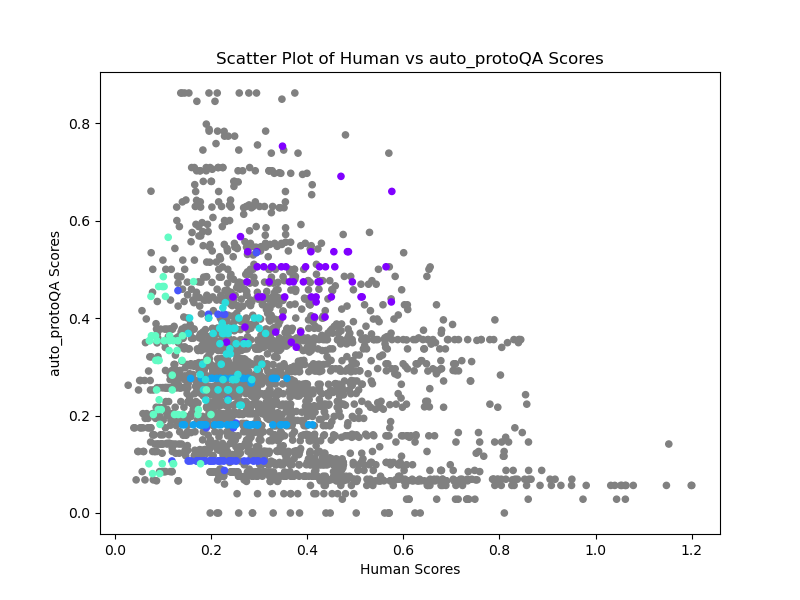}}
\subfloat[\evaluator on CFC]
{\includegraphics[width=0.238\textwidth]{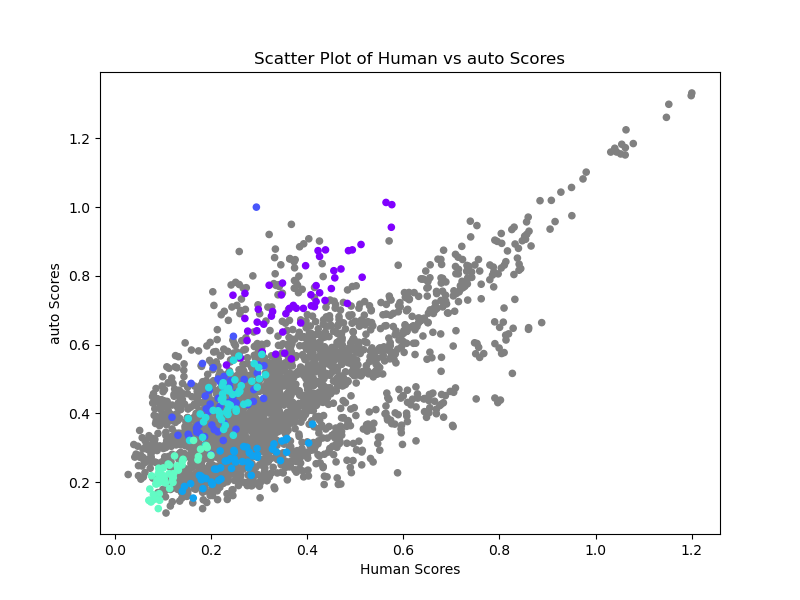}}\\
\subfloat[\small ProtoQA Score on ProtoQA]
{\includegraphics[width=0.24\textwidth]{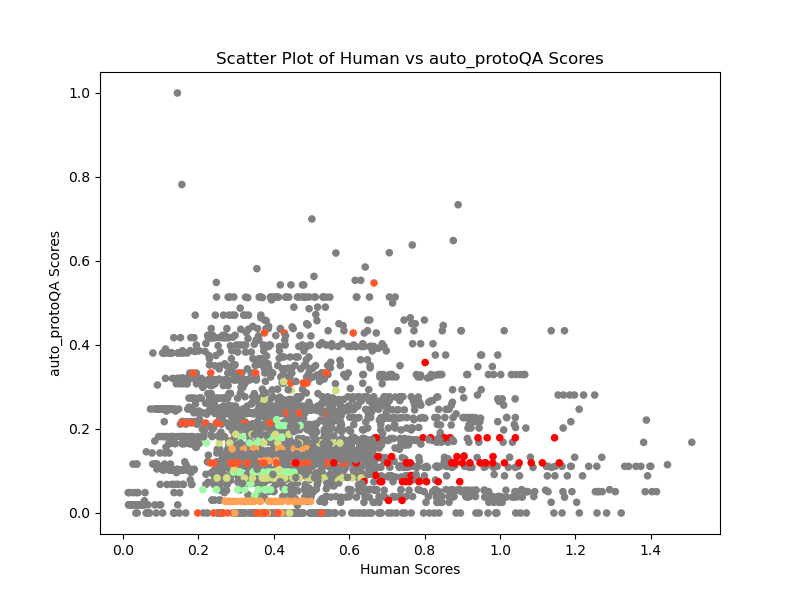}}
\subfloat[\evaluator on ProtoQA]
{\includegraphics[width=0.238\textwidth]{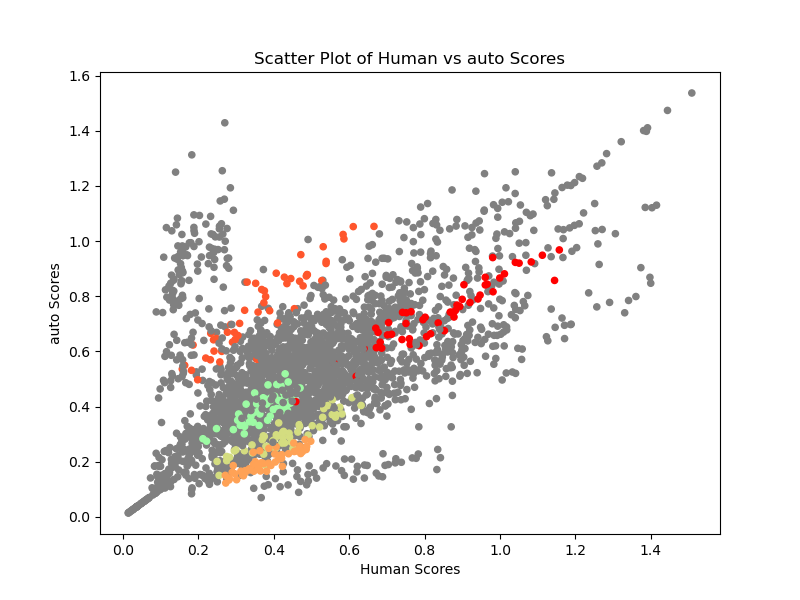} }
\caption{Scatter plots of ProtoQA evaluator and \evaluator for two datasets. The X-axis is gold KL score with human annotations, and the y-axis is automatic score with human cluster and WordNet matching. (a) and (b) show for 1-ProtoQA Max-10 score and \evaluator on CFC dev.
(c) and (d) are the same score methods on ProtoQA dev. Sampled five different questions are annotated with different colors. In both datasets, we see positive correlated trend with \evaluator, while ProtoQA evaluator barely correlates with gold scores.}
\label{fig:scatter_cfc}
\end{figure}

In order to qualitatively understand the correlation score gap between ProtoQA evaluator and \evaluator shown in table \ref{tab:correlation-dev}, we plotted KL values when using human clustering and wordnet matching function, shown in figure \ref{fig:scatter_cfc}. It is evident from figures (b) and (d) that \evaluator exhibits a clear positive correlation with the gold KL score. In contrast, the ProtoQA score tends to be a horizontal line, showing minimal correlation with the gold scores. We hypothesize that ProtoQA is less sensitive to subtle changes in the answer distribution, and we will verify in the next section.

\subsubsection{Prediction Error Types}
\begin{table}[]
\centering
\begin{tabular}{@{}cc|ccc@{}}
\toprule
                                                                              & Clustering & MA             & WR             & WS             \\ \midrule
\multirow{4}{*}{\begin{tabular}[c]{@{}c@{}}ProtoQA \\ Evaluator\end{tabular}} & Human      & 0.733          & 0.320          & -.025         \\ \cmidrule(l){2-5} 
                                                                              & Gmeans     & 0.744          & 0.465          & 0.033          \\
                                                                              & Xmeans     & 0.721          & 0.409          & 0.035          \\
                                                                              & HAC        & 0.678          & 0.4            & -.018         \\ \midrule
\multirow{4}{*}{\evaluator}                                                          & Human      & \textbf{0.875} & \textbf{0.791} & \textbf{0.245} \\ \cmidrule(l){2-5} 
                                                                              & Gmeans     & 0.745          & 0.706          & 0.187          \\
                                                                              & Xmeans     & 0.782          & 0.68           & 0.231          \\
                                                                              & HAC        & 0.711          & 0.642          & 0.188          \\ \bottomrule
\end{tabular}
\caption{Average Spearman correlation between human evaluation and automatic evaluation when model prediction includes different error types, MA - missing answer, WR - wrong ranking, and WS - wrong score for CFC dev questions. The matching function for all the entries in this table is WordNet (see Table \ref{tab:correlation-CFC-dev-new-2} for the correlation using other matching functions).
}
\label{tab:correlation-CFC-dev-new-1}
\end{table}
We designed three sampling techniques to mimic common errors in model predictions when the model is tasked to predict multiple correct answers with accurate probability calibration. These sampling methods aim to reveal the performance difference between evaluators when given varying error types ranging from easy to hard to identify. Example samplings can be seen in Figure \ref{fig:wrong score}. A list of answer examples are shown in section A.2 in the Appendix.

\paragraph{Missing Answers} The first type of error is one of the most common and easily made errors when predicting multiple correct answers, i.e missing one or more correct answers in prediction. In the implementation, we intentionally delete a random number of categorical probabilities of the ground truth distribution, $P_g$, from $P_h$. 

\paragraph{Wrong Ranking} If the model did not miss any correct answers, we ask the next question of whether it gives the correct ranking. For wrong ranking sampling, the probabilities of categories of the ground truth distribution, $P_G$, are randomly switched so that the answer ranking is wrong. 

\paragraph{Wrong Score} The last error type assumes the model predicts all correct answers with the correct ranking, however, the scores are not well calibrated. This could be the hardest task but can also be extremely important for high steak domains where the model needs to be well calibrated. For this sampling, the categorical ranking is kept the same but the categorical probabilities of the ground truth distribution, $P_G$, are varied to a random degree.

The results are shown in Table \ref{tab:correlation-CFC-dev-new-1}. For missing answers, both evaluators achieved correlation scores over 0.7, indicating a consistent alignment with human scores. However, \evaluator significantly outperforms the ProtoQA evaluator when the prediction includes incorrectly ranked responses. This demonstrates that the primary discrepancy between the ProtoQA and CFC evaluators arises from the ProtoQA evaluator's inaccurate judgment when the prediction includes wrong rankings. A particularly notable finding emerges from the wrong score errors, given this error type is extremely hard to identify, even for humans, \evaluator achieves positive correlation scores, while the ProtoQA evaluator exhibits nearly zero correlation. This performance gap can be attributed to the ProtoQA max-10's limitation of considering only the first 10 correct answers. In contrast, \evaluator considers all answers and is able to capture these finer changes, resulting in its ability to evaluate more nuanced differences between model predictions.

\section{CFC Results}
\begin{table}[]
\centering
\begin{tabular}{@{}c|cccc@{}}
\toprule
Set           & \multicolumn{2}{c|}{Dev}                           & \multicolumn{2}{c}{Test}      \\ \midrule
Shot          & ZS            & \multicolumn{1}{c|}{FS}            & ZS            & FS            \\ \midrule
GPT2-L        & 1.67          & \multicolumn{1}{c|}{1.07}          & 1.49          & 1.12          \\
GPT2-XL       & 1.32          & \multicolumn{1}{c|}{1.03}          & 1.14          & 0.85          \\
ProtoQA FT    & 0.80          & \multicolumn{1}{c|}{0.79}          & \textbf{0.61} & 0.70          \\
GPT2-L FT     & 0.76          & \multicolumn{1}{c|}{0.70}          & 0.68          & 0.71          \\
GPT 3.5 turbo & \textbf{0.66} & \multicolumn{1}{c|}{0.64}          & 0.67          & \textbf{0.61} \\
GPT 4         & 0.67          & \multicolumn{1}{c|}{\textbf{0.59}} & 0.66          & 0.68          \\
LLAMA2        & 0.85          & \multicolumn{1}{c|}{0.87}          & 0.82          & 0.85          \\
Human         & \multicolumn{2}{c|}{0.18}                          & \multicolumn{2}{c}{0.06}      \\ \bottomrule
\end{tabular}

\caption{Model performance on CFC (\textbf{lower is better}). ZS means zero-shot, and FS means one-shot prediction. GPT2-L and GPT2-XL is the GPT2 large and XL model respectively, ProtoQA FT is the ProtoQA fine-tuned, while GPT2-L FT is our own fined-tuned model. }
\label{table:cfc_model_result}
\vspace{-1em}
\end{table}
Given the high correlation of \evaluator with human gold KL scores, we employ \evaluator in evaluating CFC model performance. All the evaluator parameter are tuned on CFC dev data, then fix the parameters to report results on CFC test data.
\label{result:exp}

\paragraph{Baseline Models}
In order to generate different answers for the same prompt, we use Nucleus Sampling~\citep{holtzman2019curious}. We generate 200 sampled answers from the \gpt Large model and 100 answers for the \gpt XL model for each question and treat them as the model prediction. We experimented with temperatures from 0.1 to 1.0, and chose the model parameters with the best dev performance, then reported the test performance.

We conducted experiments using various large language models, employing Hugging Face's PyTorch GPT2 Large and XL models \cite{Wolf2019HuggingFacesTS, gpt2} and OpenAI's API for versions 3.5-turbo and 4 \cite{bian2023chatgpt, openai2023gpt4}. Our tests spanned zero-shot, one-shot, and fine-tuning scenarios using the ProtoQA dataset.

In one-shot experiments, we reformatted CFC questions as "[Q]: {context, question}, [A]" and included a sample Q\&A pair from the CFC dev set to familiarize the model with the format. For fine-tuning, we used the ProtoQA pre-trained model and also trained GPT2 Large with a similar task format for 3 epochs on an Nvidia M40 GPU, denoted as GPT2-L FT in our results.

\paragraph{Human Performance}
In order to get a human performance on this task, we collected 30 additional human responses and evaluated them the same way as a model prediction.

\paragraph{Discussion}
As shown in Table \ref{table:cfc_model_result}, the best performing model are large models, i.e GPT3.5/GPT4, or fine-tuned GPT2 model. Large models showed significant performance improvement compared to other models, even without fine-tuning data. However, all the model performances still have a large gap compared to human. This indicates the proposed benchmark combined with the probabilistic measurement \evaluator is able to identify the performance gap between LLMs and humans, and leaves us ample space to improve the model.

\section{Conclusion}
In this paper, we assert that commonsense is an implicit probability distribution over missing information, and propose a dataset that aims to evaluate commonsense in this setting via a generative question answering task; moreover, we embrace the probabilistic nature of commonsense knowledge in both the dataset creation and the metric design. We propose a probabilistic automatic evaluation \evaluator for evaluating answer distributions that is highly correlated to human judgment. Using this metric, we observe that model performance on our new dataset is significantly worse than human performance, indicating that the task is sufficiently challenging. In the future, we aim to further extend the size of the dataset, both in number of instances as well as answer length.

\section{Limitation}
We acknowledge that our collected answers are not nearly as perfect for populations around the world and coule be biased towards populations from certain regions, in this case, English speakers in the US. But we argue this framework is one step in the right direction, and we leave the collection for broader cultures for future work. 

The collected data also only included a limited scope of commonsense. We believe that extending the CFC to other commonsense domains, like temporal and social understanding, is important. For example, the social outcome of a particular event could vary significantly depending on culture, and each outcome is valid. The size of the proposed dataset is also limited; however, considering the number of annotations we collected for each question, the size is decent. We also proposed a probabilistic measurement \evaluator, which is needed in the era of LLMs to identify model limitations. 

We also acknowledge that the proposed evaluator \evaluator could be susceptible to adversarial attacks due to the automation and flexibility nature of the evaluation module. It is possible to have a model that achieves high scores but performs poorly for the task. We will explore the combination of symbolic and neural methods to increase the robustness of the evaluator in future work.

\section*{Acknowledgments}
We thank the NLP group at the University of Pittsburgh for their feedback on the early draft of the paper. We thank Joey Hou at Pitt for his help in data annotation. We also thank the anonymous reviewers for their feedback to help us improve the paper. Lorraine was partially supported by the DARPA MCS program at the beginning stage of the work. Any opinions, findings, conclusions, or recommendations expressed in this paper are those of the authors and do not necessarily reflect the sponsor.

\bibliography{anthology,custom}

\begin{thebibliography}{38}
\expandafter\ifx\csname natexlab\endcsname\relax\def\natexlab#1{#1}\fi

\bibitem[{Aydin et~al.(2014)Aydin, Yilmaz, Li, Li, Gao, and Demirbas}]{aydin2014crowdsourcing}
Bahadir~Ismail Aydin, Yavuz~Selim Yilmaz, Yaliang Li, Qi~Li, Jing Gao, and Murat Demirbas. 2014.
\newblock Crowdsourcing for multiple-choice question answering.
\newblock In \emph{AAAI}, pages 2946--2953. Citeseer.

\bibitem[{Banarescu et~al.(2013)Banarescu, Bonial, Cai, Georgescu, Griffitt, Hermjakob, Knight, Koehn, Palmer, and Schneider}]{Banarescu2013AbstractMR}
Laura Banarescu, Claire Bonial, Shu Cai, Madalina Georgescu, Kira Griffitt, Ulf Hermjakob, Kevin Knight, Philipp Koehn, Martha Palmer, and Nathan Schneider. 2013.
\newblock Abstract meaning representation for sembanking.
\newblock In \emph{LAW@ACL}.

\bibitem[{Bhagavatula et~al.(2020)Bhagavatula, Bras, Malaviya, Sakaguchi, Holtzman, Rashkin, Downey, tau Yih, and Choi}]{bhagavatula2020abductive}
Chandra Bhagavatula, Ronan~Le Bras, Chaitanya Malaviya, Keisuke Sakaguchi, Ari Holtzman, Hannah Rashkin, Doug Downey, Wen tau Yih, and Yejin Choi. 2020.
\newblock \href {https://openreview.net/forum?id=Byg1v1HKDB} {Abductive commonsense reasoning}.
\newblock In \emph{International Conference on Learning Representations}.

\bibitem[{Bian et~al.(2023)Bian, Han, Sun, Lin, Lu, and He}]{bian2023chatgpt}
Ning Bian, Xianpei Han, Le~Sun, Hongyu Lin, Yaojie Lu, and Ben He. 2023.
\newblock \href {http://arxiv.org/abs/2303.16421} {Chatgpt is a knowledgeable but inexperienced solver: An investigation of commonsense problem in large language models}.

\bibitem[{Bojanowski et~al.(2017)Bojanowski, Grave, Joulin, and Mikolov}]{fasttex}
Piotr Bojanowski, Edouard Grave, Armand Joulin, and Tomas Mikolov. 2017.
\newblock Enriching word vectors with subword information.
\newblock \emph{Transactions of the Association for Computational Linguistics}, 5:135--146.

\bibitem[{Boratko* et~al.(2020)Boratko*, Li*, O{'}Gorman*, Das*, Le, and McCallum}]{boratko2020protoqa}
Michael Boratko*, Xiang~Lorraine Li*, Tim O{'}Gorman*, Rajarshi Das*, Dan Le, and Andrew McCallum. 2020.
\newblock {P}roto{QA}: A question answering dataset for prototypical common-sense reasoning.
\newblock In \emph{Conference on Empirical Methods in Natural Language Processing, EMNLP}.

\bibitem[{Cai and Lam(2020)}]{cai-lam-2020-amr}
Deng Cai and Wai Lam. 2020.
\newblock \href {https://doi.org/10.18653/v1/2020.acl-main.119} {{AMR} parsing via graph-sequence iterative inference}.
\newblock In \emph{Proceedings of the 58th Annual Meeting of the Association for Computational Linguistics}, pages 1290--1301, Online. Association for Computational Linguistics.

\bibitem[{Chater et~al.(2006)Chater, Tenenbaum, and Yuille}]{CHATER2006287}
Nick Chater, Joshua~B. Tenenbaum, and Alan Yuille. 2006.
\newblock \href {https://doi.org/https://doi.org/10.1016/j.tics.2006.05.007} {Probabilistic models of cognition: Conceptual foundations}.
\newblock \emph{Trends in Cognitive Sciences}, 10(7):287--291.
\newblock Special issue: Probabilistic models of cognition.

\bibitem[{Chen et~al.(2023)Chen, Shi, Fu, Cheng, Li, and Xiao}]{chen2023say}
Jiangjie Chen, Wei Shi, Ziquan Fu, Sijie Cheng, Lei Li, and Yanghua Xiao. 2023.
\newblock \href {http://arxiv.org/abs/2305.05976} {Say what you mean! large language models speak too positively about negative commonsense knowledge}.

\bibitem[{Devlin et~al.(2019)Devlin, Chang, Lee, and Toutanova}]{bert}
Jacob Devlin, Ming-Wei Chang, Kenton Lee, and Kristina Toutanova. 2019.
\newblock Bert: Pre-training of deep bidirectional transformers for language understanding.
\newblock In \emph{NAACL}.

\bibitem[{Feng et~al.(2022)Feng, Zhou, Wang, Jin, and Roth}]{Feng2022GenericTR}
Yu~Feng, Ben Zhou, Haoyu Wang, Helen~Jingshu Jin, and Dan Roth. 2022.
\newblock \href {https://api.semanticscholar.org/CorpusID:254877129} {Generic temporal reasoning with differential analysis and explanation}.
\newblock In \emph{Annual Meeting of the Association for Computational Linguistics}.

\bibitem[{Harremo{\"e}s and Tusn{\'a}dy(2012)}]{harremoes2012information}
Peter Harremo{\"e}s and G{\'a}bor Tusn{\'a}dy. 2012.
\newblock Information divergence is more $\chi$ 2-distributed than the $\chi$ 2-statistics.
\newblock In \emph{2012 IEEE International Symposium on Information Theory Proceedings}, pages 533--537. IEEE.

\bibitem[{Holtzman et~al.(2019)Holtzman, Buys, Forbes, and Choi}]{holtzman2019curious}
Ari Holtzman, Jan Buys, Maxwell Forbes, and Yejin Choi. 2019.
\newblock The curious case of neural text degeneration.
\newblock \emph{arXiv preprint arXiv:1904.09751}.

\bibitem[{Huang et~al.(2019)Huang, Le~Bras, Bhagavatula, and Choi}]{huang2019cosmos}
Lifu Huang, Ronan Le~Bras, Chandra Bhagavatula, and Yejin Choi. 2019.
\newblock \href {https://doi.org/10.18653/v1/D19-1243} {Cosmos {QA}: Machine reading comprehension with contextual commonsense reasoning}.
\newblock In \emph{Proceedings of the 2019 Conference on Empirical Methods in Natural Language Processing and the 9th International Joint Conference on Natural Language Processing (EMNLP-IJCNLP)}, pages 2391--2401, Hong Kong, China. Association for Computational Linguistics.

\bibitem[{Li et~al.(2022)Li, Kuncoro, d'Autume, Blunsom, and Nematzadeh}]{li2021systematic}
Xiang~Lorraine Li, Adhi Kuncoro, Cyprien de~Masson d'Autume, Phil Blunsom, and Aida Nematzadeh. 2022.
\newblock A systematic investigation of commonsense understanding in large language models.
\newblock \emph{EMNLP}.

\bibitem[{Lin et~al.(2020)Lin, Shen, Zhou, Zhou, Bhagavatula, Choi, and Ren}]{Lin2020CommonGenAC}
Bill~Yuchen Lin, Minghan Shen, Wangchunshu Zhou, Pei Zhou, Chandra Bhagavatula, Yejin Choi, and Xiang Ren. 2020.
\newblock Commongen: A constrained text generation challenge for generative commonsense reasoning.
\newblock In \emph{Findings of the Conference on Empirical Methods in Natural Language Processing, EMNLP Findings}.

\bibitem[{Liu et~al.(2019)Liu, Ott, Goyal, Du, Joshi, Chen, Levy, Lewis, Zettlemoyer, and Stoyanov}]{roberta}
Yinhan Liu, Myle Ott, Naman Goyal, Jingfei Du, Mandar Joshi, Danqi Chen, Omer Levy, Mike Lewis, Luke Zettlemoyer, and Veselin Stoyanov. 2019.
\newblock Roberta: A robustly optimized bert pretraining approach.
\newblock \emph{ArXiv}, abs/1907.11692.

\bibitem[{Mardia et~al.(2020)Mardia, Jiao, T{\'a}nczos, Nowak, and Weissman}]{mardia2020concentration}
Jay Mardia, Jiantao Jiao, Ervin T{\'a}nczos, Robert~D Nowak, and Tsachy Weissman. 2020.
\newblock Concentration inequalities for the empirical distribution of discrete distributions: beyond the method of types.
\newblock \emph{Information and Inference: A Journal of the IMA}, 9(4):813--850.

\bibitem[{Mikolov et~al.(2013)Mikolov, Chen, Corrado, and Dean}]{word2vec}
Tomas Mikolov, Kai Chen, Gregory~S. Corrado, and Jeffrey Dean. 2013.
\newblock Efficient estimation of word representations in vector space.
\newblock In \emph{ICLR}.

\bibitem[{Moss(2018)}]{Moss2018-MOSPK-2}
Sarah Moss. 2018.
\newblock \emph{Probabilistic Knowledge}.
\newblock Oxford University Press, Oxford, United Kingdom.

\bibitem[{Murtagh and Legendre(2014)}]{murtagh2014ward}
Fionn Murtagh and Pierre Legendre. 2014.
\newblock Ward’s hierarchical agglomerative clustering method: which algorithms implement ward’s criterion?
\newblock \emph{Journal of classification}, 31(3):274--295.

\bibitem[{Novikov(2019)}]{Novikov2019}
Andrei Novikov. 2019.
\newblock \href {https://doi.org/10.21105/joss.01230} {{PyClustering}: Data mining library}.
\newblock \emph{Journal of Open Source Software}, 4(36):1230.

\bibitem[{{OpenAI}(2023)}]{openai2023gpt4}
{OpenAI}. 2023.
\newblock Introducing gpt-4.
\newblock \url{https://openai.com/blog/gpt-4/}.
\newblock Accessed: [Insert date here].

\bibitem[{Pavese(2020)}]{Pavese2020-PAVPKI}
Carlotta Pavese. 2020.
\newblock \href {https://doi.org/10.1093/analys/anz094} {Probabilistic knowledge in action}.
\newblock \emph{Analysis}, 80(2):342--356.

\bibitem[{Pelleg et~al.(2000)Pelleg, Moore et~al.}]{pelleg2000x}
Dan Pelleg, Andrew~W Moore, et~al. 2000.
\newblock X-means: Extending k-means with efficient estimation of the number of clusters.
\newblock In \emph{Icml}, volume~1, pages 727--734.

\bibitem[{Pennington et~al.(2014)Pennington, Socher, and Manning}]{glove}
Jeffrey Pennington, Richard Socher, and Christopher~D. Manning. 2014.
\newblock Glove: Global vectors for word representation.
\newblock In \emph{EMNLP}.

\bibitem[{Pillutla et~al.(2021)Pillutla, Swayamdipta, Zellers, Thickstun, Welleck, Choi, and Harchaoui}]{pillutla2021mauve}
Krishna Pillutla, Swabha Swayamdipta, Rowan Zellers, John Thickstun, Sean Welleck, Yejin Choi, and Zaid Harchaoui. 2021.
\newblock Mauve: Measuring the gap between neural text and human text using divergence frontiers.
\newblock \emph{Advances in Neural Information Processing Systems}, 34:4816--4828.

\bibitem[{Radford et~al.(2019)Radford, Wu, Child, Luan, Amodei, Sutskever et~al.}]{gpt2}
Alec Radford, Jeffrey Wu, Rewon Child, David Luan, Dario Amodei, Ilya Sutskever, et~al. 2019.
\newblock Language models are unsupervised multitask learners.
\newblock \emph{OpenAI blog}, 1(8):9.

\bibitem[{Recasens and Hovy(2011)}]{recasens2011blanc}
Marta Recasens and Eduard Hovy. 2011.
\newblock Blanc: Implementing the rand index for coreference evaluation.
\newblock \emph{Natural Language Engineering}.

\bibitem[{Sakaguchi et~al.(2020)Sakaguchi, Le~Bras, Bhagavatula, and Choi}]{winogrande}
Keisuke Sakaguchi, Ronan Le~Bras, Chandra Bhagavatula, and Yejin Choi. 2020.
\newblock Winogrande: An adversarial winograd schema challenge at scale.
\newblock In \emph{Proceedings of the AAAI Conference on Artificial Intelligence}, volume~34, pages 8732--8740.

\bibitem[{Sap et~al.(2019{\natexlab{a}})Sap, Rashkin, Chen, Le~Bras, and Choi}]{sap2019socialiqa}
Maarten Sap, Hannah Rashkin, Derek Chen, Ronan Le~Bras, and Yejin Choi. 2019{\natexlab{a}}.
\newblock \href {https://doi.org/10.18653/v1/D19-1454} {Social {IQ}a: Commonsense reasoning about social interactions}.
\newblock pages 4463--4473.

\bibitem[{Sap et~al.(2019{\natexlab{b}})Sap, Rashkin, Chen, Le~Bras, and Choi}]{sap-etal-2019-social}
Maarten Sap, Hannah Rashkin, Derek Chen, Ronan Le~Bras, and Yejin Choi. 2019{\natexlab{b}}.
\newblock \href {https://doi.org/10.18653/v1/D19-1454} {Social {IQ}a: Commonsense reasoning about social interactions}.
\newblock In \emph{Proceedings of the 2019 Conference on Empirical Methods in Natural Language Processing and the 9th International Joint Conference on Natural Language Processing (EMNLP-IJCNLP)}, pages 4463--4473, Hong Kong, China. Association for Computational Linguistics.

\bibitem[{Talmor et~al.(2019)Talmor, Herzig, Lourie, and Berant}]{talmor2018commonsenseqa}
Alon Talmor, Jonathan Herzig, Nicholas Lourie, and Jonathan Berant. 2019.
\newblock Commonsenseqa: A question answering challenge targeting commonsense knowledge.
\newblock In \emph{NAACL}.

\bibitem[{Wolf et~al.(2019)Wolf, Debut, Sanh, Chaumond, Delangue, Moi, Cistac, Rault, Louf, Funtowicz, and Brew}]{Wolf2019HuggingFacesTS}
Thomas Wolf, Lysandre Debut, Victor Sanh, Julien Chaumond, Clement Delangue, Anthony Moi, Pierric Cistac, Tim Rault, R'emi Louf, Morgan Funtowicz, and Jamie Brew. 2019.
\newblock Huggingface's transformers: State-of-the-art natural language processing.
\newblock \emph{ArXiv}, abs/1910.03771.

\bibitem[{Zellers et~al.(2018)Zellers, Bisk, Schwartz, and Choi}]{zellers2018swagaf}
Rowan Zellers, Yonatan Bisk, Roy Schwartz, and Yejin Choi. 2018.
\newblock Swag: A large-scale adversarial dataset for grounded commonsense inference.
\newblock In \emph{EMNLP}.

\bibitem[{Zellers et~al.(2019)Zellers, Holtzman, Bisk, Farhadi, and Choi}]{hellaswag}
Rowan Zellers, Ari Holtzman, Yonatan Bisk, Ali Farhadi, and Yejin Choi. 2019.
\newblock Hellaswag: Can a machine really finish your sentence?
\newblock In \emph{Annual Meeting of the Association for Computational Linguistics, ACL}.

\bibitem[{Zhao et~al.(2008)Zhao, Guo, Xu, and Ban}]{zhao2008g}
Zhonghua Zhao, Shanqing Guo, Qiuliang Xu, and Tao Ban. 2008.
\newblock G-means: A clustering algorithm for intrusion detection.
\newblock In \emph{International Conference on Neural Information Processing}, pages 563--570. Springer.

\bibitem[{Zhou et~al.(2020)Zhou, Zhang, Cui, and Huang}]{zhou2020evaluating}
Xuhui Zhou, Yue Zhang, Leyang Cui, and Dandan Huang. 2020.
\newblock Evaluating commonsense in pre-trained language models.
\newblock In \emph{Proceedings of the AAAI conference on artificial intelligence}, volume~34, pages 9733--9740.

\end{thebibliography}
\bibliographystyle{acl_natbib}

\appendix

\section{Appendix}
\label{sec:appendix}

\subsection{Validation of the CFC Evaluator}
\label{sec:validation}
\subsubsection{Diverse Sampling }
\label{sec:Diverse Sampling}

\begin{table}[]
\centering
\begin{tabular}{@{}cccc@{}}
\toprule
                                                                              & \multicolumn{1}{l}{} & \begin{tabular}[c]{@{}c@{}}ProtoQA\\  Dataset\end{tabular} & \begin{tabular}[c]{@{}c@{}}CFC \\ Dataset\end{tabular} \\ \midrule
\multirow{4}{*}{\begin{tabular}[c]{@{}c@{}}ProtoQA \\ Evaluator\end{tabular}} & Human                & 0.111                                                      & -0.082                                                 \\
                                                                              & Gmeans               & 0.471                                                      & 0.391                                                  \\
                                                                              & Xmeans               & 0.375                                                      & 0.3                                                    \\
                                                                              & HAC                  & 0.309                                                      & 0.27                                                   \\ \midrule
\multirow{4}{*}{\begin{tabular}[c]{@{}c@{}}CFC \\ Evaluator\end{tabular}}     & Human                & 0.829                                                      & \textbf{0.855}                                         \\
                                                                              & Gmeans               & \textbf{0.881}                                             & 0.773                                                  \\
                                                                              & Xmeans               & 0.718                                                      & 0.765                                                  \\
                                                                              & HAC                  & 0.748                                                      & 0.785                                                  \\ \bottomrule
\end{tabular}
\caption{Average Spearman correlation between human evaluation and automatic evaluation under diverse sampling strategy for CFC dev questions with matching function being WordNet (see Table \ref{tab:correlation-dev-auto-diverse} for the evaluation score using other matching function).
}
\label{tab:correlation-dev-diverse}
\end{table}

\begin{table}[]
\begin{tabular}{@{}llrrllllll@{}}
\toprule
\multicolumn{2}{l}{}  & \multicolumn{1}{c}{\begin{tabular}[c]{@{}c@{}}ProtoQA\\ Dataset\end{tabular}} & \multicolumn{1}{c}{\begin{tabular}[c]{@{}c@{}}CFC\\ Dataset\end{tabular}} \\ \midrule
Clustering & Matching & \multicolumn{1}{l}{Diverse}                                                   & \multicolumn{1}{l}{Diverse}                                                 \\ \cmidrule(r){1-4}
Human      & Cosine   & 0.886                                                                         & 0.754                                                                       \\
Human      & GR       & 0.891                                                                         & 0.752                                                                       \\ \cmidrule(r){1-4}
Gmeans     & Cosine   & 0.616                                                                         & 0.661                                                                       \\
Gmeans     & GR       & 0.607                                                                         & 0.682                                                                       \\
Xmeans     & Cosine   & 0.674                                                                         & 0.646                                                                       \\
Xmeans     & GR       & 0.665                                                                         & 0.646                                                                       \\
HAC        & Cosine   & 0.696                                                                         & \textbf{0.701}                                                              \\
HAC        & GR       & \textbf{0.699}                                                                & 0.673                                                                       \\ \bottomrule
\end{tabular}
\caption{Average Spearman correlation for CFC evaluator between human evaluation and automatic evaluation under diverse sampling strategy for ProtoQA dev and CFC dev questions with matching functions being human annotation, Cosine similarity function, or Gaussian Regression.
}
\label{tab:correlation-dev-auto-diverse}
\end{table}

We employed a naive sampling stragegy called diverse sampling to emulate the a noisy scenario where the prediction set is  a linear combination of the ground-truth distribution $\hat{P}_g$ and a uniform distribution $\hat{P}_u$. The formulation of diverse sampling is $p = \alpha \hat{P}_g + (1 - \alpha)\hat{P}_u$ where $\alpha$ is drawn from a uniform distribution between $0$ and $1$. The comparison result between CFC evaluator and protoQA evaluator using diverse sampling are shown in Table \ref{tab:correlation-dev-diverse} and comparison of CFC evaluator under different automatic clustering and matching are in Table \ref{tab:correlation-dev-auto-diverse}.

\subsection{Examples of sampling methods}
\paragraph{Question}: A helicopter is being used to tackle the disaster. Who will use helicopters to tackle the disaster?

\paragraph{Ground Truth Answers} A selection of ground truth answers and their probability are:
\begin{itemize}
  \item Cluster 1: “pilot”, p (C1) = 0.5
  \item Cluster 2: "disaster management team", p(C2) = 0.3
  \item Cluster 3: "people" p(C3) = 0.2.
\end{itemize}
In terms of probabilistic distribution, p(cluster 1) > p(cluster 2) > p(cluster 3) for the ground truth.

\paragraph{Wrong Score Sampling} The distribution could be: 
\begin{itemize}
  \item Cluster 1: “pilot”, p(C1) = 0.7
  \item Cluster 2: "disaster management team", p(C2) = 0.2
  \item Cluster 3: "people", p(C3) = 0.1
\end{itemize}
The ranking of distribution is kept the same but the actual probabilities of the clusters are much different than that of the ground truth clusters.

\paragraph{Wrong Ranking Sampling} The distribution could be:
\begin{itemize}
  \item Cluster 1: “pilot”, p(C1) = 0.2
  \item Cluster 2: "disaster management team", p(C2) = 0.3
  \item Cluster 3: "people" p(C3) = 0.5
\end{itemize}
The ranking of distribution is now cluster 3 > cluster 2 > cluster 1 which is different from the ground truth ranking.

\paragraph{missing answer sampling} The distribution could be:
\begin{itemize}
    \item Cluster 1: “pilot”, p(C1)=0.7
    \item Cluster 3: "people", p(C3)=0.3.
\end{itemize}
Note that Cluster 2 (disaster management team) is missing from the sampled distribution.

\begin{figure*}[ht]
\centering
\includegraphics[width=0.32\textwidth]{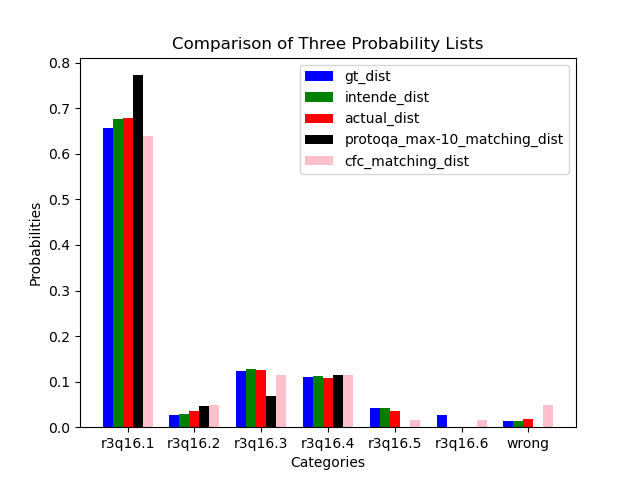}
\includegraphics[width=0.32\textwidth]{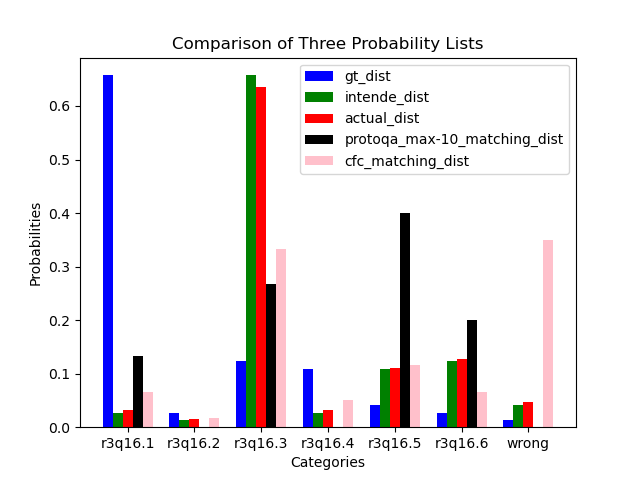}
\includegraphics[width=0.32\textwidth]{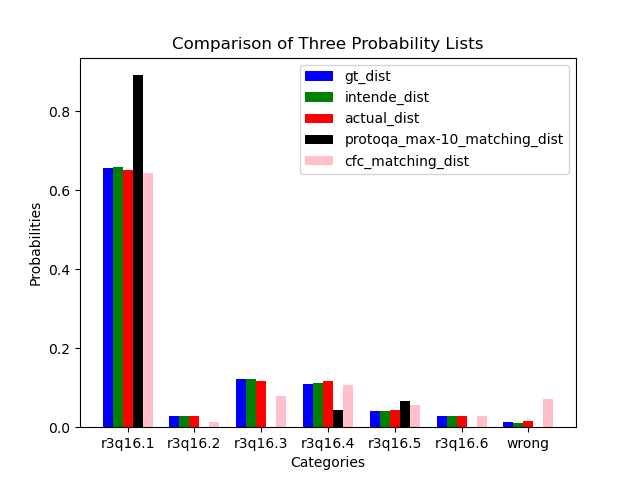}
\caption{Example sampling step for missing answer sampling (left), wrong ranking sampling (middle), and wrong score sampling (right).}
\label{fig:wrong score}
\end{figure*}

\begin{figure*}[h]
\centering
\small
{\includegraphics[width=0.43\textwidth]{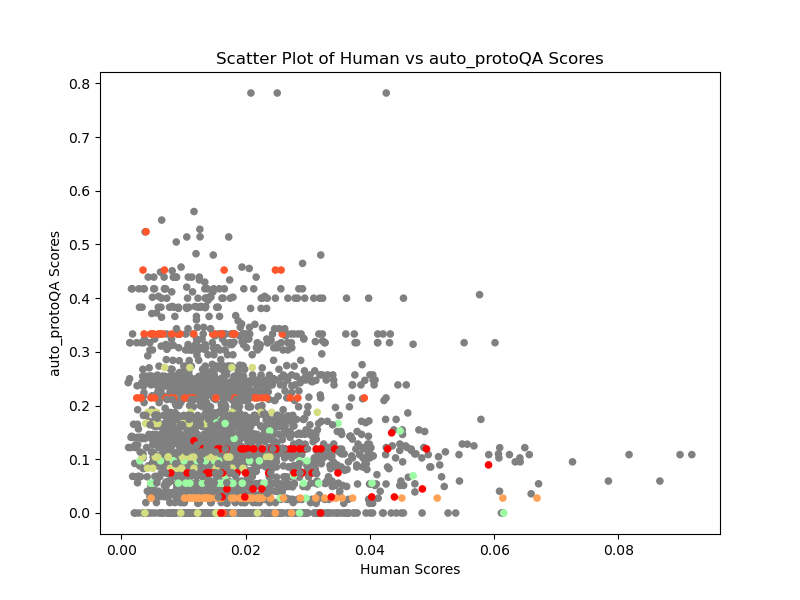}}
{\includegraphics[width=0.43\textwidth]{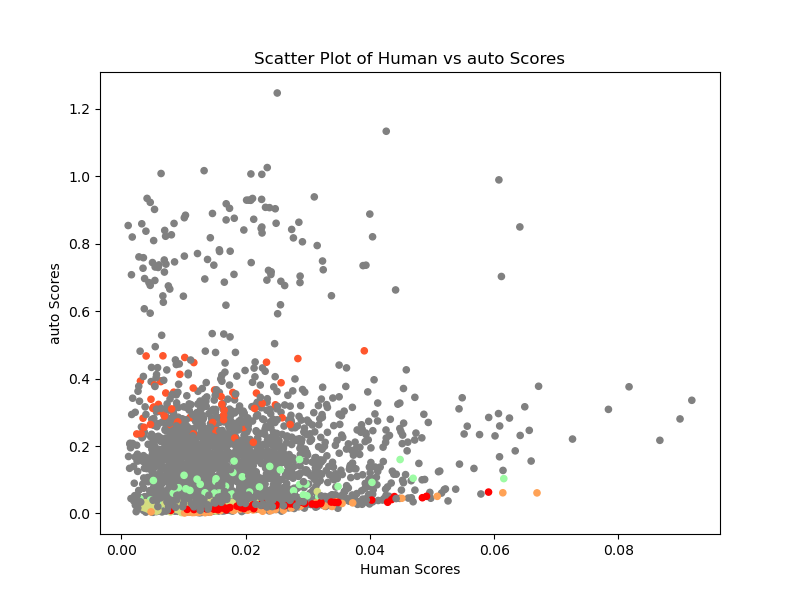} }
  \caption{Comparison of correlation between ProtoQA evaluator and CFC evaluator for Wrong Score sampled questions in ProtoQA with ground-truth clusters. The X-axis is the evaluaotr score with human assignment, and the y-axis is the KL value with WordNet assignment. The figure on the left is the correlation for ProtoQA Max-10 with human clustering and WordNet matching. The figure on the right is the correlation for CFC evaluator with human clustering and WordNet matching. These corresponds to the ProtoQA Evaluator / Human / WordNet row and the CFC Evaluator / Human / WordNet row with column being WS in Table \ref{tab:correlation-CFC-dev-new-1}. Different questions are annotated with different colors.}
\label{fig:scatter_protoQA_wrong_score}
\end{figure*}

\begin{figure*}[t]
    \centering
    \includegraphics[width=\textwidth]{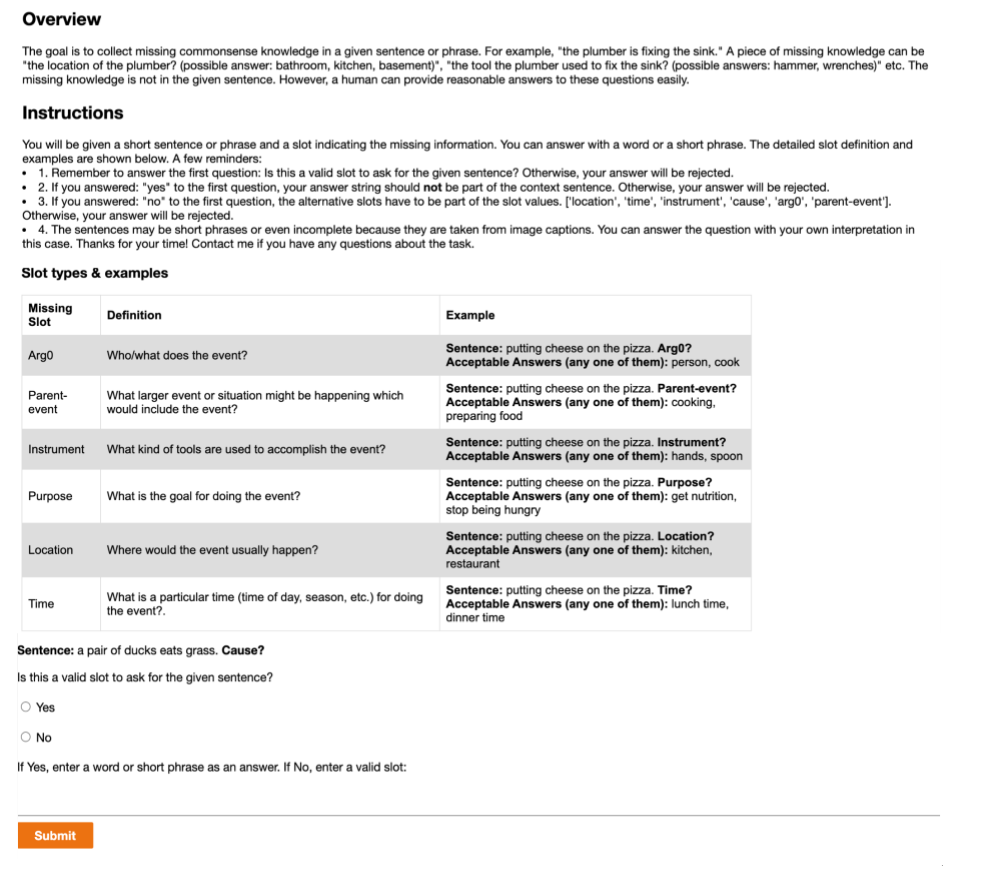}
    \caption{The screen shot for the the dataset collection page in Amazon MTurk.}
    \label{fig:AMT}
\end{figure*}

\begin{table}[]
\begin{tabular}{@{}ccccc@{}}
\toprule
Clustering & Matching & MA             & WR             & WS             \\ \midrule
Human      & Cosine   & 0.740          & 0.501          & 0.141          \\
Human      & GR       & 0.765          & 0.499          & 0.125          \\ \midrule
Gmeans     & Cosine   & 0.763          & \textbf{0.599} & 0.086          \\
Gmeans     & GR       & \textbf{0.787} & 0.556          & 0.051          \\
Xmeans     & Cosine   & 0.773          & 0.519          & 0.090          \\
Xmeans     & GR       & 0.759          & 0.504          & 0.096          \\
HAC        & Cosine   & 0.694          & 0.593          & 0.143          \\
HAC        & GR       & 0.698          & 0.580          & \textbf{0.162} \\ \bottomrule
\end{tabular}
\caption{Average Spearman correlation between human evaluation and automatic evaluation under MA - missing answer, WR - wrong ranking, and WS - wrong score sampling strategies for CFC dev questions with matching functions being human annotation, Cosine similarity function, or Gaussian Regression.
}
\label{tab:correlation-CFC-dev-new-2}
\end{table}

\end{document}